\documentclass[letterpaper, 10 pt, conference]{ieeeconf}  % Comment this line out if you need a4paper

\IEEEoverridecommandlockouts                              % This command is only needed if 
\usepackage{amsmath}
\usepackage{graphicx}
\usepackage{amssymb}
\usepackage{float}
\usepackage{booktabs}
\usepackage{algorithmicx}
\usepackage{microtype}
\usepackage{times}
\usepackage{graphicx}
\usepackage{cancel}
\usepackage[space]{cite}
\usepackage{pdfsync}
\usepackage{balance}
\usepackage{color}
\usepackage{algpseudocode}
\usepackage{mathtools}
\usepackage{multirow}
\usepackage{makecell}
\usepackage{bm}
\usepackage{changepage}
\usepackage{algorithm}

\usepackage{diagbox}
\usepackage{epstopdf}
\usepackage{pifont}
\usepackage{multirow}
\usepackage{url}
\usepackage{tabularx}
\usepackage[linkcolor=black,citecolor=black,urlcolor=black,colorlinks=true]{hyperref}
\usepackage{verbatim} 
\usepackage[skip=3pt, font={small}]{caption}

\title{\LARGE \bf
HGS-Planner: Hierarchical Planning Framework for Active Scene Reconstruction Using 3D Gaussian Splatting}

\author{Zijun Xu$^{1}$, Rui Jin$^{2}$, Ke Wu$^{1}$, Yi Zhao$^{1}$, Zhiwei Zhang$^{1}$, Jieru Zhao$^{3}$,  Fei Gao$^{2}$\\ Zhongxue Gan$^{1}$ and Wenchao Ding$^{1}$$^{*}$ % <-this % stops a space
\thanks{$^{1}$Zijun Xu,  Ke Wu, Yi Zhao,  Zhiwei Zhang, Zhongxue Gan,  Wenchao Ding are with the Academy for Engineering \& Technology, Fudan University, Shanghai, China, 200433. 
{\tt\small \{zijunxu23, kewu23, yizhao24, zhiweizhang23\}@m.fudan.edu.cn, \{ganzhongxue, dingwenchao\}@fudan.edu.cn}}
\thanks{$^{2}$Rui Jin, Fei Gao are with the Institute of Cyber-Systems and Control, College of Control Science and Engineering, Zhejiang University, Hangzhou 310027, China. 
        {\tt\small bbbbigrui@gmail.com, fgaoaa@zju.edu.cn}}%
\thanks{$^{3}$Jieru Zhao is with the Department of Computer Science and Engineering, Shanghai Jiao Tong University
        {\tt\small zhao-jieru@sjtu.edu.cn}}
        \thanks{Corresponding author: Wenchao Ding}%
}

\makeatletter
\let\@oldmaketitle\@maketitle% Store \@maketitle
\renewcommand{\@maketitle}{\@oldmaketitle
	\vspace{0.2cm}
	\centering
	\setcounter{figure}{0}
	\begin{minipage}{1.0\linewidth}
		\includegraphics[width=1.0\textwidth]{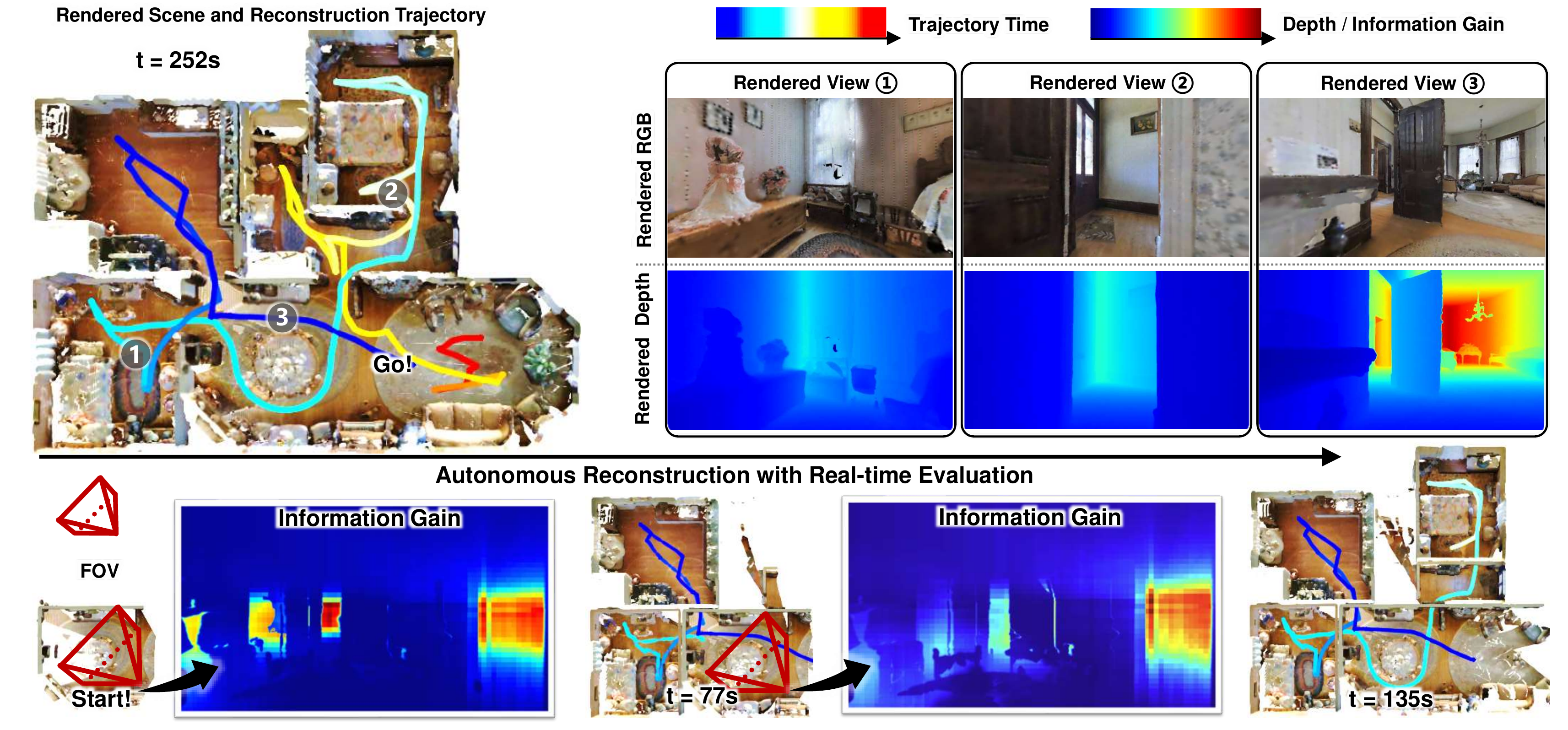}
		\vspace{0cm}
		\captionof{figure}{\label{fig:simu_res}In a simulated complex house scene, we implemented our high-fidelity active reconstruction system on a mobile robot equipped with an RGB-D sensor. The colored curves represent the robot's executed trajectories. We showcase the reconstruction results, which include the entire rendered scene, detailed renderings from three different views, and the variation in information gain at a specific view.}
	\end{minipage}
	\vspace{-0.3cm}
}

\makeatother

\begin{document}

\maketitle
\thispagestyle{empty}
\pagestyle{empty}

% \begin{figure*}[t]
% \centering
% \includegraphics[width=1\linewidth]{teaser111.png}
% \caption{In a simulated complex house scene, we implemented our high-fidelity active reconstruction system on a mobile robot with an RGB-D sensor. The colored curves in the visuals represent the robot's executed trajectories. We showcase the reconstruction results, which include the entire rendered scene, detailed renderings from three different views, and the variation in information gained at a specific view.}
% \label{fig.:1}
% \end{figure*}

%%%%%%%%%%%%%%%%%%%%%%%%%%%%%%%%%%%%%%%%%%%%%%%%%%%%%%%%%%%%%%%%%%%%%%%%%%%%%%%%
\begin{abstract}
In complex missions such as search and rescue, robots must make intelligent decisions in unknown environments, relying on their ability to perceive and understand their surroundings. High-quality and real-time reconstruction enhances situational awareness and is crucial for intelligent robotics. Traditional methods often struggle with poor scene representation or are too slow for real-time use. Inspired by the efficacy of 3D Gaussian Splatting (3DGS), we propose a hierarchical planning framework for fast and high-fidelity active reconstruction. Our method evaluates completion and quality gain to adaptively guide reconstruction, integrating global and local planning for efficiency. Experiments in simulated and real-world environments show our approach outperforms existing real-time methods.
\end{abstract}
\section{INTRODUCTION}

In tasks such as search and rescue or target finding, which rely on active exploration, robots must preserve as much geometric and texture information from the environment as possible to support effective decision-making \cite{arnold2018search, rs15133266}.  Online active reconstruction plays a crucial role in these missions by enabling robots to construct and update environmental models in real time, allowing them to navigate and adapt more efficiently in complex and unknown environments. 

However, conventional active reconstruction\cite{isler2016information, zhou2021fuel, 9812330} that fuse sensor data across space and time only capture coarse structures and struggle with rich scene details and novel view evaluation. Recently, Neural Radiance Field (NeRF)\cite{mildenhall2021nerf}-based methods \cite{ran2023neurar,zeng2023efficient,zeng2024autonomous, Wu_Zhang_Gao_Zhao_Gan_Ding_2024} have gained popularity for their high-fidelity scene representation and efficient memory usage. However, NeRF’s inherent volumetric rendering process requires dense sampling of every pixel, resulting in long training times and poor real-time performance\cite{wu2024hgsmappingonlinedensemapping}. Additionally, its use of implicit neural representations makes it challenging to evaluate reconstruction quality accurately in real time. In fact, active reconstruction systems demand quick responses and the ability to make decisions based on real-time reconstruction quality dynamically. NeRF’s computational bottlenecks make it unsuitable for scene representation in active reconstruction, especially in scenarios that require real-time responses.

Compared to NeRF, 3D Gaussian Splatting (3DGS)\cite{kerbl20233d} offers a more efficient explicit representation,  reducing computational complexity and better suiting online active reconstruction\cite{jin2024gsplannergaussiansplattingbasedplanningframework}. Additionally, the Gaussian map's real-time integration of new data gives it the potential to provide immediate feedback on the rendering quality of new views. However, despite these notable advantages, the application of 3DGS for active reconstruction in unknown environments is still largely unexplored. 

Though high-quality reconstruction can be achieved using 3DGS, the task with 3D Gaussian representation faces three main challenges. First, efficiently and accurately evaluating novel view quality without ground truth is crucial for guiding robot motion planning, but it remains challenging. Second, while efficiency is critical to active reconstruction, Gaussian maps can only represent occupied areas, posing a challenge for efficiently reconstructing unobserved regions. Third, effectively integrating Gaussian map data into closed-loop motion planning is essential for active reconstruction, yet how to do this effectively is still an open question. 

To address the above problems, we propose an efficient 3D Gaussian-based real-time planning framework for active reconstruction. To the best of our knowledge, our framework is the pioneering work exploring 3DGS representation for online active reconstruction. Firstly, we introduce Fisher Information, which represents the expectation of observation information and is independent of ground truth\cite{Jiang2023FisherRF}, to evaluate novel view quality gain in online reconstruction. Secondly, we improve exploration efficiency in 3D Gaussian representation by integrating unknown voxels into the splatting-based rendering process, allowing us to assess new viewpoints' coverage of unexplored areas. Thirdly, we use Gaussian map data to adaptively select viewpoints, which balances reconstruction quality and efficiency, and integrate it into an active planning framework. Our experimental results confirm that our framework supports efficient and high-quality online reconstruction.

To summarize, our contributions are:
\begin{itemize}

    \item To the best of our knowledge, we propose the first online adaptive hierarchical autonomous reconstruction system using 3DGS.

    \item We design a novel viewpoint selection strategy based on reconstruction coverage and quality and implement it within an autonomous reconstruction framework.

    \item We conduct extensive simulation and real-scene experiments to validate the effectiveness of the proposed system.
\end{itemize}

\begin{figure*}
    \centering
    \includegraphics[width=\linewidth]{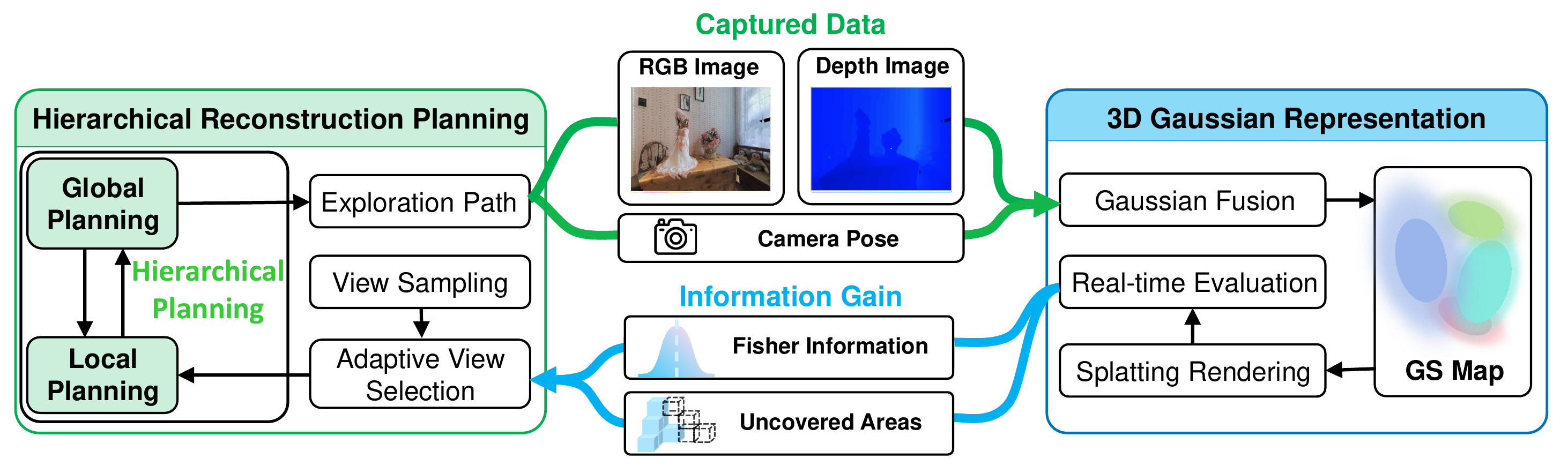}
\caption{An overview of our efficient autonomous reconstruction system with high-fidelity. Utilizing 3DGS for scene representation, the unobserved areas and the Fisher Information from the GS map are provided in real-time to evaluate the quality and completeness of the online reconstruction. Our proposed active reconstruction planning framework efficiently guides the robot to acquire new scene data, ensuring a comprehensive and
high-fidelity 3DGS reconstruction.}
\vspace{-0.5cm}
\label{pipeline}
\end{figure*}

\section{RELATED WORK}

\subsection{High-fidelity Reconstruction Representation}

% For high-fidelity reconstruction, various scene representations are used, including meshes, planes, and surfel clouds. Recently, Neural Radiance Fields (NeRF) \cite{mildenhall2021nerf} have gained prominence due to their photorealistic rendering capabilities. NeRF methods can be categorized into three types: implicit, hybrid, and explicit.

Various scene representations are used for reconstruction, including meshes, planes, and surfel clouds. Recently, Neural Radiance Field (NeRF) \cite{mildenhall2021nerf} has gained prominence due to its photorealistic rendering capabilities. NeRF methods can be categorized into three types: implicit, hybrid representation, and explicit. Implicit method \cite{sucar2021imap} is memory-efficient but faces challenges such as catastrophic forgetting and significant computational overhead in larger scenes. Hybrid representation methods\cite{zhu2022nice,tie2024o2v,jiang2023h2} integrate the benefits of implicit MLPs with structural features, significantly improving scene scalability and precision. The explicit method introduced in \cite{fridovich2022plenoxels} directly embeds map features within voxels, bypassing the use of MLPs, which allows for faster optimization. 

Although NeRF excels in photorealistic reconstruction \cite{ran2023neurar}, 
its ray sampling approach leads to high computational costs, making it impractical for real-time autonomous reconstruction\cite{jin2024gsplannergaussiansplattingbasedplanningframework}. In contrast, 3DGS\cite{kerbl20233d} facilitates real-time rendering of novel views through its fully explicit representation and innovative differential splatting rendering, which has been utilized in real-time SLAM, allowing the scene reconstruction from RGB-D images\cite{keetha2023splatam, yan2023gs,wu2024hgsmappingonlinedensemapping}. 
\subsection{Active Reconstruction System}
 The active reconstruction system integrates data acquisition into the decision-making loop, guiding robots in data collection tasks\cite{jin2024gsplannergaussiansplattingbasedplanningframework}. Scene representations can categorize these systems: voxel-based methods\cite{zhou2021fuel,schmid2020efficient,9812330}, surface-based methods\cite{huang2018active,gao2022meeting,schmid2020efficient}, neural network-based methods\cite{Feng_2024_CVPR,ran2023neurar} and 3D Gaussian-based methods\cite{jin2024gsplannergaussiansplattingbasedplanningframework}. 

Voxel methods\cite{zhou2021fuel,schmid2020efficient, 9812330} use compact grids for efficient space representation, while surface-based methods\cite{huang2018active,gao2022meeting,schmid2020efficient} focus on geometric details. However, both largely neglect color and texture details. Neural network-based methods, such as NeurAR\cite{ran2023neurar} and Naruto\cite{Feng_2024_CVPR}, combine NeRF with Bayesian models for view planning but are computationally intensive, causing frequent delays. 3D Gaussian Splatting (3DGS) offers high-fidelity scene representation and fast data fusion, but its application in active reconstruction is still rare. GS-Planner \cite{jin2024gsplannergaussiansplattingbasedplanningframework} combines 3DGS with voxel maps but lacks effective information gain evaluation and relies on random sampling, reducing efficiency and risking local optima.

\section{METHOD}

\subsection{Problem Statement and System Overview}
This study aims to efficiently explore unknown and spatially constrained 3D environments and reconstruct high-quality 3D models using a mobile robot by generating a trajectory composed of a sequence of paths and viewpoints\cite{ran2023neurar}. In previous greedy-based NBV methods\cite{Feng_2024_CVPR,zeng2023efficient}, the path design seeks to identify the trajectory leading to the next optimal view. However, from a global perspective, this approach always converges on local optima, reducing reconstruction efficiency significantly. We design a hierarchical autonomous reconstruction framework through a novel viewpoint selection criterion, selecting a series of optimal viewpoints for global and local path planning, enabling rapid and high-fidelity reconstruction. 
As illustrated in Fig. \ref{pipeline}, our proposed hierarchical autonomous reconstruction framework consists of two main components. \textbf{The 3D Gaussian Representation} module reconstructs high-fidelity scenes and offers real-time evaluations of potential future viewpoints by leveraging 3DGS’s efficient data fusion and online rendering capabilities. These evaluations encompass gains in both coverage information and reconstruction quality. \textbf{The Active Reconstruction Planning} module is divided into two subcomponents: global planning and local planning. Global planning generates a path that enhances exploration efficiency and avoids local optima, while local planning identifies optimal viewpoints through view sampling and adaptive selection, developing a local path. Finally, the global and local paths are merged into an exploration path that guides the robot's movement. 

\subsection{3D Gaussian Representation} 
We use SplaTam \cite{keetha2024splatam}, a 3D Gaussian-based SLAM method, for online 3D Gaussian Splatting reconstruction. The scene is represented as numerous isotropic 3D Gaussian, each characterized by eight parameters: center position $\xi  \in \mathbb{R}^{3}$, RGB color  $r \in \mathbb{R}^{3}$, radius $\mu \in \mathbb{R}$, and opacity $\rho \in \mathbb{R}$. The opacity function $\pi$ of a point $\alpha \in \mathbb{R}^{3}$, computed from each 3D Gaussian, is defined as follows:
\begin{equation}
    \pi (\alpha, \rho) = \rho \exp \left(-\frac{|\alpha-\xi|^{2}}{2 \mu^{2}}\right).
\end{equation}
We adopt a differentiable approach to render the images to optimize the Gaussian parameters for scene representation. The final rendered RGB color $R_{pix}$ and depth $D_{pix}$ can be mathematically formulated as the alpha blending of N sequentially ordered points that overlap the pixel,
\begin{equation}
\begin{aligned}
    R_{pix} &= \sum_{i=1}^{N} r_{i} \pi_{i} \prod_{j=1}^{i-1}\left(1-\pi_{j}\right), \\
    D_{pix} &= \sum_{i=1}^{N} d_{i} \pi_{i} \prod_{j=1}^{i-1}\left(1-\pi_{j}\right).
\end{aligned}
\end{equation}
where $ d_{i}$ is the depth of the \textit{i}-th 3D Gaussian center, corresponding to the z-coordinate of its center position in the camera coordinate system.
\subsection{Reconstruction Coverage Gain Evaluation \label{sentionc}}

To improve the efficiency and completeness of scene reconstruction, we implemented an evaluation of reconstruction coverage gain for candidate viewpoints. Calculating the increase in reconstructed areas from a new viewpoint requires considering occupied and unobserved regions. Yet, under the 3D Gaussian representation, we can only recognize the former, making it difficult to determine which regions have yet to be observed. To solve this problem, similar to GS-Planner\cite{jin2024gsplannergaussiansplattingbasedplanningframework},  we maintain a voxel map to represent unobserved volume and integrate it into the splatting rendering. However, unlike GS-Planner,  we employ a more streamlined calculation method that leverages uniform voxel volumes to achieve model-consistent pixel-level reconstruction coverage gain within the 3DGS rendering process.
\begin{figure}[t]
    \begin{center}
        \includegraphics[width=0.9\columnwidth]{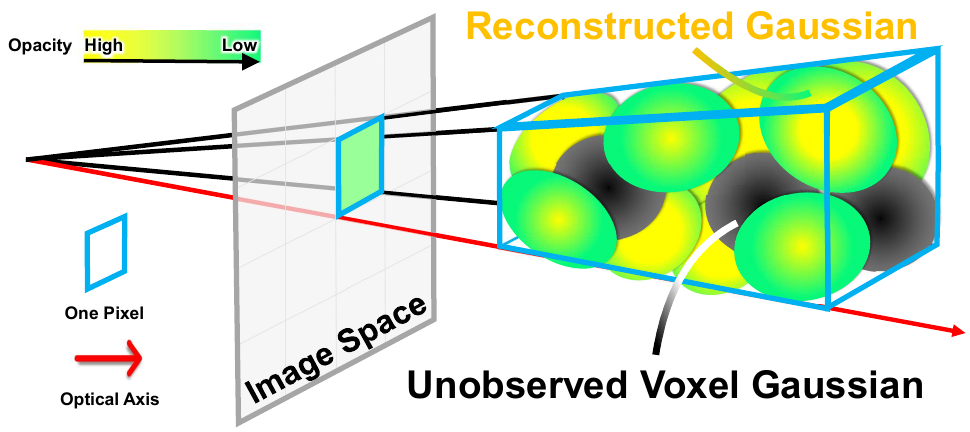}
    \end{center}
    \caption{A 3D illustration of pixel-level coverage gain evaluation. Given a set of reconstructed Gaussians and a viewpoint, the coverage gain is rendered by weighting unobserved voxel Gaussians with the transmittance of reconstructed Gaussians along the optical ray. }
    \label{fig:coverage_gain}
    \vspace{-0.9cm}
\end{figure}
Specifically, given a set of 3D Gaussians and a viewpoint pose, we first sort the Gaussians from front to back by depth. Then, using the ordered 3D Gaussians, we can efficiently render depth images by alpha-compositing the splatted 2D projection of each Gaussian sequentially in pixel space. During rendering, by integrating the unobserved voxels from the maintained voxel map into the Gaussian map, we can determine whether an unobserved region exists between adjacent Gaussians. Considering both the uniform volume of each voxel and the inherent opacity attribute of the Gaussians, we can evaluate the visibility gain of unobserved regions  for each viewpoint by utilizing a transmittance weight, which can be expressed as:
\begin{equation}
\begin{aligned}V_{pix}=\sum_{i=1}^nV\prod_{j=1}^{m_i}{(1-\alpha_j)}\end{aligned}.
\end{equation}
where n is the number of unobserved volumes along the ray, $m_i$ is the number of the related 3D Gaussians before the \textbf{i}-th unobserved voxel Gaussian, $\prod_{j=1}^{m_i}{(1-\alpha_j)}$ is the transmittance weight, V represents the same unobserved voxel volume.
Leveraging the fast splatting-based rendering, the Reconstruction Coverage evaluation process runs in parallel with the reconstruction process, resulting in highly efficient overall computation. To illustrate the coverage evaluation process more intuitively, we provide Fig.  \ref{fig:coverage_gain}.
\subsection{Reconstruction Quality Gain Evaluation}
To enhance reconstruction quality and accuracy, we employ Fisher Information to quantify the quality gains from novel viewpoints, leveraging its independence from ground truth\cite{Jiang2023FisherRF}.
\begin{figure}[t]
    \begin{center}
        \includegraphics[width=0.9\columnwidth]{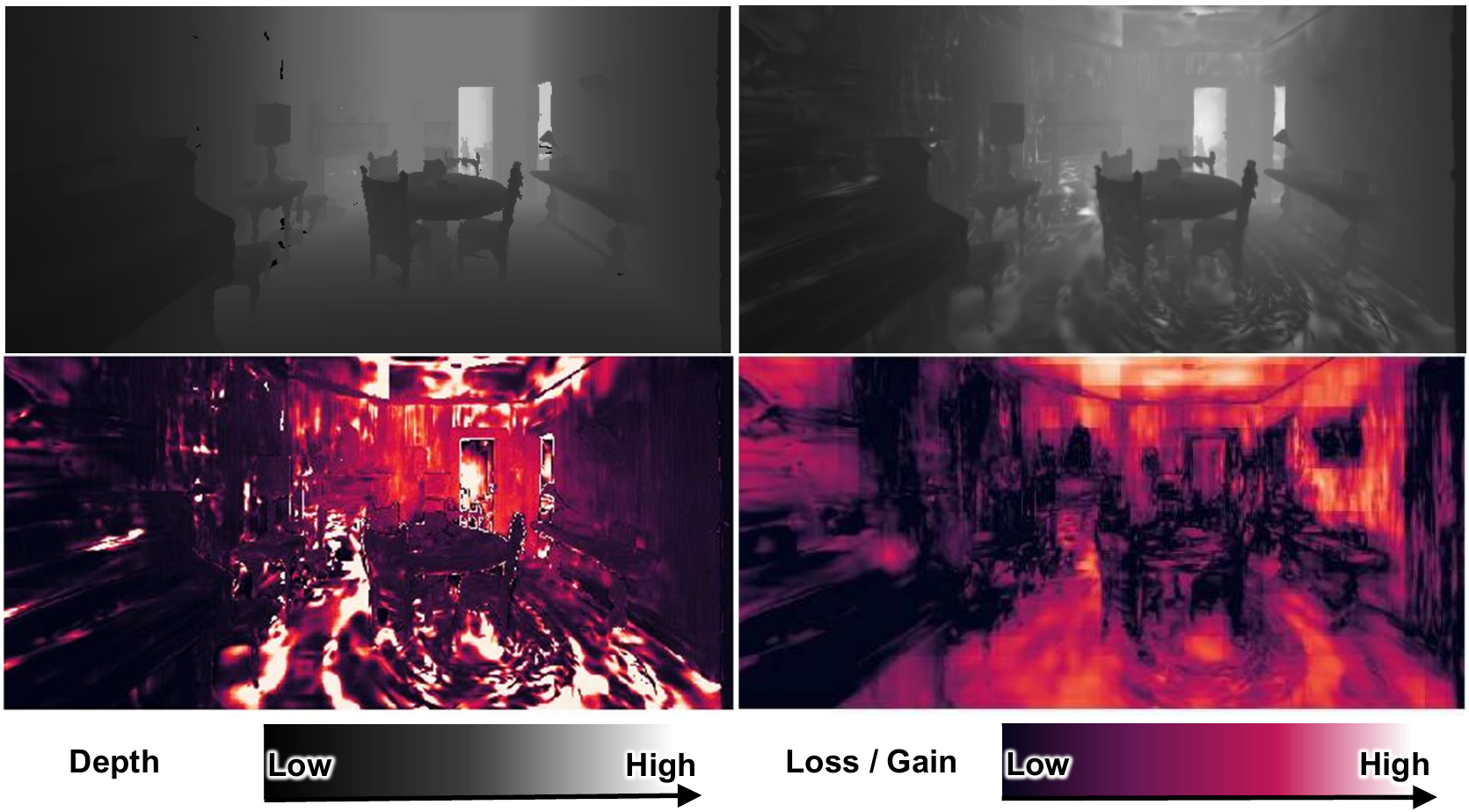}
    \end{center}
    \caption{The top images are depth maps: the left is the ground truth, and the right is the rendered depth. Below, the left image shows the squared error, and the right illustrates the quality gain. }
    \label{fig:quality_gain}
    \vspace{-1.cm}
\end{figure}
The primary goal of  neural rendering is to minimize the negative log-likelihood (NLL) between rendered and ground truth images, described by:
\begin{equation}\label{logp}
-\log\mathbb{P}(\boldsymbol{\Psi}|\mathbf{x},\mathbf{w})=\left(\boldsymbol{\Psi}-f(\mathbf{x},\mathbf{w})\right)^{T}\left(\boldsymbol{\Psi}-f(\mathbf{x},\mathbf{w})\right).
\end{equation}
where $\mathbf{x}$ is the camera pose, $\Psi$ the corresponding image, $\mathbf{w}$ the model parameters, and $f(\mathbf{x},\mathbf{w})$ the rendering model. Under the regularity conditions\cite{schervish2012theory}, Fisher Information for Eq. \ref{logp} is defined as the Hessian of the log-likelihood function concerning $\mathbf{w}$:
\begin{equation}
    \mathcal{I}(\mathbf{w})=-\mathbb{E}_{\mathbb{P}(\boldsymbol{\Psi}|\mathbf{x},\mathbf{w})}\left[\frac{\partial^2\log \mathbb{P}(\boldsymbol{\Psi}|\mathbf{x},\mathbf{w})}{\partial\mathbf{w}^2}\bigg|\mathbf{w}\right]=\mathbf{H}^{\prime\prime}[\boldsymbol{\Psi}|\mathbf{x},\mathbf{w}],
\end{equation}
where $\mathbf{H}^{\prime\prime}[\boldsymbol{\Psi}|\mathbf{x},\mathbf{w}]$ is the Hessian of Eq. \ref{logp}. In the evaluation process, we can obtain the initial estimation of parameters $\mathbf{w}^{*}$ by using $\{\mathbf{\Psi}_{i}^{acq}\}$ as the training set $D^{train}$. Our quality evaluation purpose is to identify the viewpoints that can maximize the Information Gain\cite{lindleyEIG,kirsch2022unifying,HoulsbyBAL} among the viewpoints $\mathbf{x}_{i}^{acq}\ \in D^{candidate} $ in comparison to $D^{train}$, where $D^{candidate}$ represents the collection of candidate viewpoints:
\begin{equation}
 \begin{aligned}\label{entropy-acq}
&\mathcal{I}[\mathbf{w}^{*};\{\mathbf{\Psi}_{i}^{acq}\}|\{\mathbf{x}_{i}^{acq}\},D^{train}] \\
&=H[\mathbf{w}^{*}|D^{train}]-H[\mathbf{w}^{*}|\{\mathbf{\Psi}_{i}^{acq}\},\{\mathbf{x}_{i}^{acq}\},D^{train}],
\end{aligned}
\end{equation} where $H[\cdot]$ is the entropy\cite{kirsch2022unifying}.
Considering the log-likelihood form in Eq. \ref{logp}, specifically the rendering loss, the entropy difference in the R.H.S. of Eq. \ref{entropy-acq} only depends on $H[\mathbf{w}^{*}|\{\mathbf{\Psi}_{i}^{acq}\},\{\mathbf{x}_{i}^{acq}\},D^{train}]$, then the Hessian can be approximated using just the Jacobian matrix of $f(\mathbf{x},\mathbf{w})$\cite{Jiang2023FisherRF}:
\begin{equation}\label{to_trace}
\mathbf{H}''[\mathbf{\Psi}|\mathbf{x},\mathbf{w}^*]=\nabla_\mathbf{w}f(\mathbf{x};\mathbf{w}^*)^T\nabla_\mathbf{w}f(\mathbf{x};\mathbf{w}^*).
\end{equation} As expected, the trace of Eq. \ref{to_trace} can be computed without ground truths $\{\Psi_i^{acq}\}$, as Fisher Information is independent of observations. Furthermore, with the Laplace approximation\cite{bayesian-interpolation,laplace2021}, Eq. \ref{to_trace} can be approximated by considering only diagonal elements and adding a log-prior regularizer $\lambda I$:
\begin{equation}
\mathbf{H}''[\mathbf{\Phi}|\mathbf{x},\mathbf{w}^*]\approx\mathrm{diag}(\nabla_\mathbf{w}f(\mathbf{x},\mathbf{w}^*)^T\nabla_\mathbf{w}f(\mathbf{x},\mathbf{w}^*))+\lambda I.
\end{equation} Like coverage reconstruction, we integrate quality evaluation into splatting-based rendering for computational efficiency.
\subsection{Adaptive Hierarchical Planning}
To avoid local optima in the exploration path, inspired by TARE\cite{9812330}, we propose an adaptive hierarchical planning framework, which combines global planning with adaptive local planning to improve the efficiency of scene reconstruction. The entire scene is divided into two regions: the local space $\mathcal{C}$ for local planning and the space outside $\mathcal{C}$ which is partitioned into evenly cuboid subspaces for global planning. 
\subsubsection{Global planning}
Each cuboid subspace is classified into three states based on the voxel map mentioned in Sec. \ref{sentionc}: "reconstructed" (only observed voxels), "reconstructing" (both observed and unobserved voxels), and "unreconstructed" (only unobserved voxels). In global planning, only "reconstructing" subspaces are taken into account. The global is to find a global path $\Gamma _{\mathrm{global}}$ that traverses all "reconstructing " subspaces, connecting their centers and the robot's current location. To achieve this, similar to \cite{9812330}, we construct a sparse random roadmap in the traversable space expanded from the past trajectory. Then we apply A* search on the roadmap to find the shortest paths among the subspaces and the current pose followed by solving a Traveling Salesman Problem (TSP)\cite{doi:10.1137/0221030} to get $\Gamma _{\mathrm{global}}$.
\subsubsection{Adaptive local planning}
Due to the trade-off between efficiency and efficacy in reconstruction, we design adaptive local planning to adjust the weights of these two aspects dynamically. Similar to the global planning approach, we use the A* algorithm combined with a TSP solver to perform local path planning after selecting the best views. The whole best views selecting algorithm is listed as  Alg.~\ref{alg1}. 

Specifically, we first calculate the intersection points between the global path and the local horizon and uniformly sample viewpoints within the local region (Lines 1-2). Then, we combine these intersections and sampled points and assess a comprehensive 360-degree information gain for each (Line 3-4). This information gain comprises two components: coverage gain and quality gain, which are weighted relative to the proportion of observed areas within the local region:
\begin{equation}
    G=\mathrm{G}(C)+\lambda_{o} \mathrm{~G}_{}(Q)
\end{equation}
where $G$ is the final information gain, $\mathrm{G}(C)$ is the coverage gain, $\mathrm{~G}_{}(Q)$ is the quality gain, $\lambda_{o}$ is the proportion of observed voxels within the local region. Subsequently, leveraging the 360-degree information gain, we select viewpoints that exceed a threshold of information gain and use a sliding-window technique to identify the optimal yaw angles(Lines 5-12). Finally, we obtain the exploration path by connecting the global and local paths. The reconstruction is complete when all cuboid subspaces are "reconstructed" and no more viewpoints are selected.
\begin{algorithm}[H]
    \caption{Adaptive local views selection}
    \label{alg1}
    \begin{algorithmic}[1]
    \Require Global Path $\Gamma_{\mathrm{global}}$, Local Horizon $\mathcal{L}$, Current Pose $\mathbf{P}_{C}$, Gaussian Map $\mathbf{M}_{G}$, Voxel map $\mathbf{M}_{V}$
    \State  Intersection Points $\mathbf{P}_{I} \leftarrow \text{CalIntersection}(\Gamma_{\mathrm{global}}, \mathcal{L})$
    \State  Sampling Points $\mathbf{P}_{S} \leftarrow \text{SamplingViewpoints}(\mathbf{P}_{C})$
    \State $\mathbf{P}_{ALL} = \mathbf{P}_{S} \cup \mathbf{P}_{I}$
    \State Gain $\mathbf{G}_{ALL} \leftarrow \text{AdaptiveEvaluation}( \mathbf{P}_{ALL})$
    
    \For{$(g_i, p_i) \in (\mathbf{G}_{ALL}, \mathbf{P}_{ALL})$}
        \If {$g_i < g_{\text{thres}}$ \textbf{and} $p_i \in \mathbf{P}_{S}$}
            \State $\mathbf{G}_{ALL} \leftarrow \mathbf{G}_{ALL} \setminus g_i$
            \State $\mathbf{P}_{ALL} \leftarrow \mathbf{P}_{ALL} \setminus p_i$
            \State \textbf{continue}
        \EndIf
    \EndFor
    \State Best Views $\mathbf{V}_{b} \leftarrow \text{SelectBestYaws}(\mathbf{G}_{ALL}, \mathbf{P}_{ALL})$
    \State Result local views: $\mathbf{V}_{b}$
    \State \textbf{Return} $\mathbf{V}_{b}$
    \end{algorithmic}
\end{algorithm}

\section{RESULTS}
\begin{table*}[!ht]
    \vspace{1.5mm}
    \centering
    \caption{Evaluations of the effectiveness and efficiency with 3DGS representation.}
    \resizebox{\textwidth}{25 mm}{
    \normalsize
    \setlength{\tabcolsep}{2mm}{
    % \scriptsize
    \begin{tabular}{c|>{\centering\arraybackslash}p{0.02\linewidth}>{\centering\arraybackslash}p{0.02\linewidth}>{\centering\arraybackslash}p{0.02\linewidth}>{\centering\arraybackslash}p{0.04\linewidth}|ccc|ccc|ccc}
    \toprule
         \multicolumn{1}{c}{}& \multicolumn{4}{c}{Variant} &\multicolumn{3}{c}{Scene\_17DRP}&\multicolumn{3}{c}{Scene\_2t7WU}&\multicolumn{3}{c}{Scene\_Gdvg}\\ 
        Method& $C_{ove.}$& $Q_{ua.}$& $H_{ier.}$& Adap& Acc↓ (cm)& Comp↓ (cm)& C.R.↑ & Acc↓ (cm)& Comp↓ (cm)& C.R.↑ & Acc↓ (cm)& Comp↓ (cm)& C.R.↑ \\ 
        \midrule
        
        V1(TARE)& & &  \checkmark&   & 2.86& 57.23& 0.47& 3.61& 28.40& 0.65& 3.03& 15.51&  0.71\\ 
        V2(Coverage)& \checkmark& &  \checkmark&  & 2.84& 7.16& 0.81& 3.21& 10.07& 0.79& 3.01& 12.97& 0.79\\
        V3(Quality)& &  \checkmark& \checkmark &  & 2.82& 54.67& 0.49& 3.19& 22.13& 0.69& 2.94& 11.27& 0.75\\
        V4(On both)& \checkmark & \checkmark & \checkmark &   & 2.81& 6.52& 0.85& 3.12& 9.63& 0.82& 2.98& 9.85& 0.85\\
        V5(Ours full) & \checkmark & \checkmark & \checkmark & \checkmark  & \textbf{2.80}& \textbf{2.66}& \textbf{0.90}& \textbf{3.09}& \textbf{2.63}& \textbf{0.91}& \textbf{1.97}& \textbf{2.63}& \textbf{0.90}\\  
     
        \midrule
         Variant & $C_{ov.}$& $Q_{ua.}$& $H_{ier.}$& Adap& $T_{VE}$ (s)& $T_{P}$ (s)& P.L. (m)& $T_{VE}$ (s)& $T_{P}$ (s)& P.L. (m)& $T_{VE}$ (s)& $T_{P}$ (s)& P.L. (m)\\ \midrule
        V1(TARE)
& & &  \checkmark&  
& 0.075& 0.098& 57.71& 0.091& 0.115& 25.97& 0.096& 0.119& 26.15\\
        V2(Coverage)
& \checkmark& &  \checkmark&  
& 0.054& 0.077& 83.27& 0.065& 0.088& 48.76& 0.059& 0.081& 27.33\\
        V3(Quality)
& &  \checkmark& \checkmark & 
& 0.052& 0.075& 60.23& 0.057& 0.082& 30.05& 0.049& 0.069& 26.46\\
        V4(On both)
& \checkmark & \checkmark & \checkmark &  
& 0.103& 0.126& 98.09& 0.118& 0.144& 58.62& 0.116& 0.137& 28.17\\
        V5(Ours full) & \checkmark & \checkmark & \checkmark & \checkmark  & 0.105& 0.129& 90.35& 0.126& 0.151& 60.52& 0.122& 0.142& 30.15\\
     
         \midrule
    \end{tabular} 
    } 
    }
    \label{table_effect_effici}
     \vspace{-0.1cm}
\end{table*}

\begin{table*}[ht]
    \vspace{-2mm}
    \centering
    \caption{Evaluations of the effectiveness and efficiency with existing planning methods.}
    \resizebox{\textwidth}{11mm}{
    \normalsize
    \setlength{\tabcolsep}{1.0mm}{
    % \scriptsize
    \begin{tabular}{c|ccccc|ccccc|ccccc}
    \toprule
        \multicolumn{1}{c}{} &\multicolumn{5}{c}{Scene\_17DRP}&\multicolumn{5}{c}{Scene\_2t7WU}&\multicolumn{5}{c}{Scene\_Gdvg}\\ 
        Method  & Acc↓ (cm)& Comp↓ (cm)& C.R.↑ & $T_{P}$ (s)& P.L. (m)& Acc↓ (cm)& Comp↓ (cm)& C.R.↑ & $T_{P}$ (s)& P.L. (m)& Acc↓ (cm)& Comp↓ (cm)& C.R.↑& $T_{P}$ (s)& P.L. (m)\\ 
        \midrule
        GS-Planner& 2.88& 4.06& 0.84& 0.147& 95.04& 3.19& 2.85& 0.88& 0.179& 74.98& 3.09& 4.76& 0.89& 0.161& 30.78\\
        
         NARUTO& 10.29& 2.87& 0.89& 0.368& 120.66& 23.94& 7.38& 0.69& 0.349& 108.01& 6.97& 4.61& \textbf{0.91}& 0.315& 90.33\\ 
         
         Ours& \textbf{2.80}& \textbf{2.66}& \textbf{0.90}& \textbf{0.129}& \textbf{90.35}& \textbf{3.09}& \textbf{2.63}& \textbf{0.91}& \textbf{0.151}& \textbf{60.52}& \textbf{1.97}& \textbf{2.63}& 0.90& \textbf{0.142}& \textbf{30.15}\\
        \midrule
    \end{tabular}
    }
    }
    \label{compare_with_existing_methods}
    \vspace{-4mm}
\end{table*}

\subsection{Implementation details}
 We run our active reconstruction system on a desktop PC with a 2.9 GHz Intel i7-10700 CPU and an NVIDIA RTX 3090 GPU, using the Autonomous Exploration Development Environment\cite{DBLP:journals/corr/abs-2110-14573} for simulation. The system's car is equipped with an RGB-D sensor and a Lidar VLP-16, providing real-time RGB-D images at 1200 $\times$680 resolution with a 5-meter range, and uses LOAM\cite{Zhang2014LOAMLO} for localization. The maximum velocity limit is 1.0 $\mathrm{~m} / \mathrm{s}$, and the depth data includes a uniform noise of 2 cm. 
 
We validate our method through simulations in three complex Matterport3D (MP3D)\cite{Matterport3D} scenes:  17DRP, 2t7WU, and Gdvg with the local planner range of 6 m $\times$ 6 m and the resolution of voxel map integrated into Gaussian map to 0.1 m. Viewpoints are sampled with a minimum distance of 1.5 m to avoid excessive overlap.

 Similar to \cite{zeng2023efficient} and \cite{ran2023neurar}, we evaluate our method in terms of effectiveness and efficiency. We adopt scene quality metrics from NARUTO\cite{Feng_2024_CVPR}: Accuracy (cm), Completion (cm), and Completion Ratio (the percentage of points in the reconstructed mesh with Completion under 5 cm). We extract geometric centroids from Gaussian spheres to simulate mesh vertices due to the absence of a standard method for converting 3DGS into mesh. In these metrics, about 300k points are sampled from the surfaces. For efficiency, similar to \cite{zeng2024autonomous}, we evaluate each step planning time (second) $T_{P}$ and the path length (meter) P.L.. For each planning cycle during the reconstruction, $T_{P}$ is divided into viewpoints sampling and evaluation time $T_{VE}$, local path planning time $T_{LP}$ and global planning time $T_{GP}$, with average $T_{GP}$ times of approximately 0.017 s (17DRP), 0.018 s (2t7WU) and 0.0 15s (Gdvg). $\mathrm{i.e.  }  T_{P}=T_{VE}+T_{LP}+T_{GP}.$ In TARE, the time taken to evaluate viewpoints corresponds to the time required to update the information about the areas they cover.
\begin{figure}[t]
    \begin{center}
        \includegraphics[width=0.95\columnwidth]{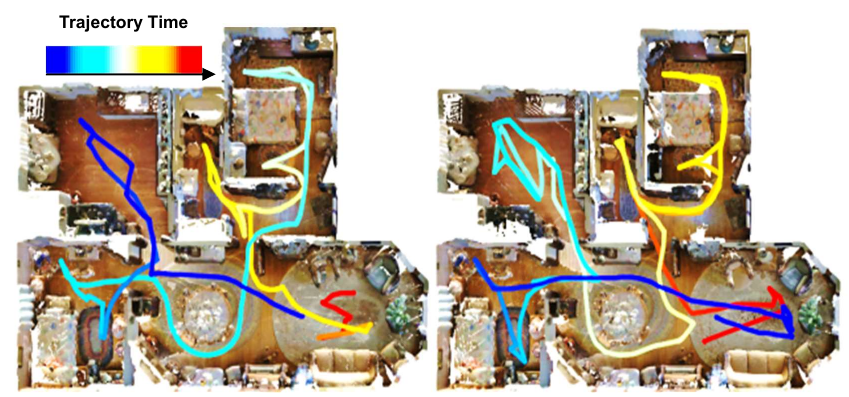}
    \end{center}
    \caption{Trajectories and the reconstruction results from the top view. Left: Ours, Right: GS-Planner}
    \label{fig:traj}
    \vspace{-1.5cm}
\end{figure}

\subsection{Efficacy of the Method}
Following \cite{zeng2024autonomous}, we evaluate our method's efficacy based on its validity and efficiency. We create variants of our method with 3D Gaussian representation: V1 (TARE \cite{9812330}), V2 (Coverage evaluation only), V3 (Quality evaluation only), and V4 (On both without an adaptive strategy). Our method proves highly effective and more efficient than other approaches.

\begin{figure*}

    \centering
    \includegraphics[width=\linewidth]{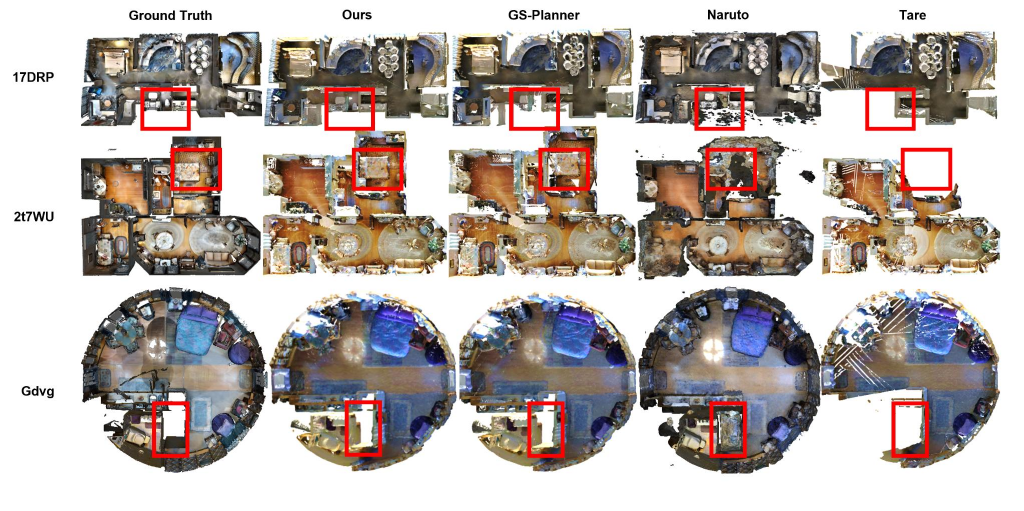}
    \vspace{-0.4cm}
    \caption{Comparison of the reconstruction scenes with different methods}
    \vspace{-0.5cm}
    \label{fig:pic_results}
\end{figure*}

1) Quality evaluation in real-time reconstruction: Fig. \ref{fig:quality_gain} shows that quality evaluation closely matches actual losses, even without ground truth. Highlighted areas on the loss map indicate regions with lower reconstruction quality, aligning with our quality gain evaluation.

2) Novel view evaluation criterion: We make V1 as a baseline applying hierarchical planning, V2 (for coverage), V3 (for quality), and V4 (on both) to verify our evaluation criterion's efficacy. Metrics in Table  \ref{table_effect_effici}  show that evaluating coverage and quality improves reconstruction. However, V2 results in low-quality reconstruction as it overlooks complex details, while V3 yields poor completeness by only refining already-covered areas and neglecting unobserved regions. 

3) Adaptive hierarchical planning: To validate the adaptive hierarchical planning, we establish V4 as our baseline. Combining these two tasks noticeably hampers the speed of scene exploration and may result in local optima, especially when dealing with intricate reconstruction details. The introduction of adaptive hierarchical planning (Ours) ensures efficient exploration while maintaining reconstruction quality, preventing the process from getting stuck in local optima. 

\subsection{Comparison with existing reconstruction methods}
We benchmark two recent works: NARUTO \cite{Feng_2024_CVPR} based on view information gain fields, and GS-Planner \cite{jin2024gsplannergaussiansplattingbasedplanningframework} using 3D Gaussian reconstruction. The metrics in Table \ref{compare_with_existing_methods} show our framework outperforms both planning efficiency and reconstruction quality. NARUTO neglects uncovered areas reducing the exploration efficiency, while GS-Planner often gets stuck in local optima during exploration. Fig. \ref{fig:pic_results} and metrics in Table \ref{compare_with_existing_methods} highlight our method's superior reconstruction. We refer readers to the supplementary video for more visual results and the reconstruction process. We implemented the GS-Planner algorithm on a mobile vehicle. Fig. \ref{fig:traj} compares the trajectories of our method and GS-Planner in scene 2t7WU, showing GS-Planner's focus on smaller areas reduces overall efficiency. Our framework achieves more efficient scene reconstruction. 

\subsection{Robot experiments in real scene}
We implemented our proposed framework on an UGV equipped with Realsense Depth Camera D435i and Ouster Lidar to perform the real scene reconstruction. FAST-LIO\cite{DBLP:journals/corr/abs-2010-08196} provides the localization. Since we use an Ackermann-steering vehicle, we replace the A* algorithm with Kino-A* to ensure the path meets kinematic constraints. The detailed process will be shown in the supplementary video.

\section{CONCLUSIONS}

In this paper, we developed a hierarchical planning framework for efficient and high-fidelity active reconstruction with 3DGS. We introduced Fisher Information to evaluate reconstruction quality and assessed coverage gain by integrating the Voxel and Gaussian maps. We also designed a novel viewpoint selection strategy within hierarchical planning. Extensive experiments show our method's superior performance. For future work, we aim to extend our research to swarm robotics in large-scale scenes.


%%%%%%%%%%%%%%%%%%%%%%%%%%%%%%%%%%%%%%%%%%%%%%%%%%%%%%%%%%%%%%%%%%%%%%%%%%%%%%%%

%%%%%%%%%%%%%%%%%%%%%%%%%%%%%%%%%%%%%%%%%%%%%%%%%%%%%%%%%%%%%%%%%%%%%%%%%%%%%%%%



% \section*{ACKNOWLEDGMENT}

% The preferred spelling of the word ÒacknowledgmentÓ in America is without an ÒeÓ after the ÒgÓ. Avoid the stilted expression, ÒOne of us (R. B. G.) thanks . . .Ó  Instead, try ÒR. B. G. thanksÓ. Put sponsor acknowledgments in the unnumbered footnote on the first page.

% %%%%%%%%%%%%%%%%%%%%%%%%%%%%%%%%%%%%%%%%%%%%%%%%%%%%%%%%%%%%%%%%%%%%%%%%%%%%%%%%

% References are important to the reader; therefore, each citation must be complete and correct. If at all possible, references should be commonly available publications.

% \begin{thebibliography}{99}

% \end{thebibliography}

\bibliographystyle{unsrt}
\bibliography{ICRA2022}

\end{document}